\documentclass[conference,compsoc]{IEEEtran}
\usepackage{blindtext, graphicx}

\ifCLASSINFOpdf

\usepackage[tight,footnotesize]{subfigure}
\usepackage{textcomp}
\usepackage{float}
\usepackage{graphicx}
\usepackage{epstopdf}
\usepackage{color}
\usepackage{wrapfig,lipsum}
\usepackage{setspace}
\usepackage{graphics}
\usepackage{grffile}
\usepackage{lipsum,multicol}
\usepackage{tabularx}
\usepackage{array}
\usepackage{algorithmic}
\usepackage{algorithm}
\usepackage{lineno}
\usepackage{mathptmx}
\usepackage{latexsym}
\usepackage{moreverb}
\usepackage{amsmath}
\usepackage{calc}
\usepackage{amssymb}
\usepackage{sidecap}
\usepackage{verbatim}
\usepackage{longtable}
\usepackage{multirow}
\usepackage{color,soul}
\usepackage{multirow}
\usepackage{microtype}
\usepackage{enumitem}

\definecolor{darkred}{RGB}{200, 50, 50}
\usepackage[style=ieee,minbibnames=1,maxbibnames=2,doi=false,isbn=false,url=false,eprint=false,date=year,bibencoding=utf8]{biblatex}
\AtEveryBibitem{\clearfield{pages}}
\AtEveryBibitem{\clearfield{volume}}
\AtEveryBibitem{\clearfield{number}}
\AtEveryBibitem{\clearfield{note}}
\AtEveryBibitem{\clearlist{language}}

\IEEEoverridecommandlockouts
\newcolumntype{C}[1]{>{\centering\let\newline\\\arraybackslash\hspace{0pt}}m{#1}}
\newcommand{\cmmnt}[1]{\ignorespaces}
\begin{document}
	
	%\bstctlcite{url.control}
	\title{
	\vspace{-0.6cm}
	{\fontsize{18.6}{23}\selectfont Importance of methodological choices in data manipulation for validating epileptic seizure detection models}
	\vspace{-0.8cm}
	
	\thanks{This work has been partially supported by the ML-Edge Swiss National Science Foundation (NSF) Research project (GA No. 200020182009/1), and the PEDESITE Swiss NSF Sinergia project (GA No. SCRSII5 193813/1). T. Teijeiro is supported by the grant RYC2021-032853-I funded by MCIN/AEI/ 10.13039/501100011033 and by the "European Union NextGenerationEU/PRTR"}
	}
	
% 	\begin{comment}
	
% 	\vspace{-2cm}
% 	\author{\IEEEauthorblockN{William Andrew Simon\IEEEauthorrefmark{1}, Una Pale\IEEEauthorrefmark{1}, Tomas Teijeiro\IEEEauthorrefmark{2}, David Atienza\IEEEauthorrefmark{1}}
% 		\IEEEauthorblockA{\IEEEauthorrefmark{1}Embedded Systems Laboratory (ESL), Swiss Federal Institute of Technology Lausanne (EPFL), Switzerland\\ \IEEEauthorrefmark{2}
% 			\{william.simon, una.pale, tomas.teijeiro, david.atienza\}@epfl.ch}
		
% 		\vspace{-0.8cm}
% 		}
	
	\author{
	\IEEEauthorblockN{Una Pale\IEEEauthorrefmark{1}, Tomas Teijeiro\IEEEauthorrefmark{1}\IEEEauthorrefmark{2}, David Atienza\IEEEauthorrefmark{1}}
	\IEEEauthorblockA{\IEEEauthorrefmark{1}Embedded Systems Laboratory (ESL), Ecole Polytechnique Federale de Lausanne (EPFL), Switzerland\\ \IEEEauthorrefmark{2} BCAM - Basque Center for Applied Mathematics, Spain \\ 
		\{una.pale, david.atienza\}@epfl.ch, tteijeiro@bcamath.org}
	
		\vspace{-0.8cm}
		}

	\maketitle

	\begin{abstract}
		
    Epilepsy is a chronic neurological disorder that affects a significant portion of the human population and imposes serious risks in the daily life of patients. Despite advances in machine learning and IoT, small, nonstigmatizing wearable devices for continuous monitoring and detection in outpatient environments are not yet available. Part of the reason is the complexity of epilepsy itself, including highly imbalanced data, multimodal nature, and very subject-specific signatures. However, another problem is the heterogeneity of methodological approaches in research, leading to slower progress, difficulty comparing results, and low reproducibility. Therefore, this article identifies a wide range of methodological decisions that must be made and reported when training and evaluating the performance of epilepsy detection systems. We characterize the influence of individual choices using a typical ensemble random-forest model and the publicly available CHB-MIT database, providing a broader picture of each decision and giving good-practice recommendations, based on our experience, where possible.

	\end{abstract}
	
	
	\IEEEpeerreviewmaketitle

	\vspace{-4mm}
	\section{Introduction}
	\label{sec:intro}
    \vspace{-2mm}
	In recent years, advances in signal processing, machine learning algorithms, the Internet of Things (IoT), and wearable devices have enabled a variety of continuous monitoring applications in many domains, particularly health monitoring.  
    One such example is epilepsy detection, with the ultimate goal of having small, non-stigmatizing, wearable devices for long-term epilepsy monitoring in patients' homes and everyday life, rather than limited to in-hospital monitoring. Epilepsy is a chronic neurological disorder characterized by the unexpected occurrence of seizures, imposing serious health risks and many restrictions on daily life. It affects a significant portion of the world's population (0.6 to 0.8\%)~\cite{mormann_seizure_2007}, of which one third of patients still suffer from seizures despite pharmacological treatments~\cite{schmidt_evidence-based_2012}.     
    Thus, there is a clear need for solutions that allow continuous unobstructed monitoring and reliable detection (and ideally prediction) of seizures~\cite{patel_patient-centered_2016, van_de_vel_automated_2016}. Furthermore, these solutions will be instrumental in the design of new treatments, assisting patients in their daily lives, and preventing possible accidents. 
    This need is also evident in the growing number of publications on seizure detection methods~\cite{rasheed_machine_2021, jory_safe_2016} and wearable devices~\cite{brinkmann_seizure_2021,verdru_wearable_2020}.  

    However, although many studies report impressive levels of accuracy via machine learning (ML) methods, the widespread adoption of commercial technology has not yet occurred. The reasons for this are many and include the specificities of epilepsy itself. For example, to properly characterize epileptic seizures, recordings must be continuous, often lasting days and leading to extremely unbalanced datasets. This imbalance must be taken into account when preparing the data set, splitting it into training and testing, training epilepsy detection models, and reporting final performance values.     
    Another challenge is the fact that epilepsy is a holistic phenomenon affecting many signal modalities, and thus, to get a full picture, multimodal data are needed from several different sensors. How to efficiently process all this data and fuse information and predictions remains an open research topic~\cite{schulze-bonhage_seizure_2022}. 
    Finally, seizures show highly personalized patterns, which require new methods of personalizing general models (that were developed from many subjects) using the characteristics of individual patients.~\cite{cogan_personalization_2016, de_cooman_personalizing_2020}. 
    
    The last reason for slower progress is that the way studies are designed, algorithms assessed, and results reported is very heterogeneous. It can be difficult to understand the level of evidence these studies provide~\cite{beniczky_standards_2018} and it is also impossible to fairly compare the results. For example, it is very difficult to compare the performance of various systems when only two quantitative values are reported (e.g., sensitivity and specificity) and when the prior probabilities vary significantly (the a priori probability of a seizure is very low, which means that the assessment of background events dominates the error calculations)~\cite{ziyabari_objective_2019}.
    
    Thus, in this paper, we want to bring attention to a number of methodological choices which are usually underreported but ultimately can have a strong influence on system performance. These choices are necessarily made during data preparation, training, and also evaluation and reporting of the results. %It is important to note that these choices and values of various parameters are not usually detailed in the papers, and without this information the reproducibility of results is also brought into question. 
    	
	The contributions of this work are summarized as follows:
	\begin{itemize}
		\item We identify a wide range of methodological decisions that must be made and reported when training and evaluating the performance of epilepsy detection systems.  
        \item We characterize and assess the influence of individual choices using a typical ensemble random-forest model and the publicly available CHB-MIT database.
		\item We provide a broader picture of each decision and give good-practice recommendations where possible. 
	\end{itemize}
	
	The remainder of the paper is organized as follows: Section~\ref{sec:choices} details the relevant methodological choices and their potential influence. Section~\ref{sec:Expsetup} provides a description of the experimental setup used to evaluate the influence of these methodological choices and parameters. Section~\ref{sec:ExpResults} presents the experimental results, while Section~\ref{sec:Discusion} comments on more broad and general observations on the presented results. It also presents certain methodological recommendations for the development of future epileptic seizure detection algorithms, as well as more general time series analysis applications. Section~\ref{sec:Conclusion} concludes this work.
	
	%%%%%%%%%%%%%%%%%%%%%%%%%%%%%%%%%%%%%%%%%%%%%%%%%%%%%%%%%%%%%%%%%%%%%%%%%%%%%%%%%%%%%%%%%%%%%%%%%%%%%%%%%%%%%%%%%%%%%%%%%%%%%%%%%%%%%%%%%%%%%%%%
    \vspace{-2mm}
	\section{Methodological choices} 
	\label{sec:choices}
    \vspace{-2mm}
    There are many methodological choices to make when evaluating machine learning algorithms and systems in terms of their performance and suitability for real-life applications. These choices, %even though sometimes seeming small and irrelevant, 
    can significantly impact the performance and repeatability of such results in practice. 
    %Yet, we often overlook some of them, go with the most common ones, or forget about the possibility of testing them. 
    In this section we go through the most important choices, discussing data preparation, training and testing methodology, and performance measures, as listed in Table~\ref{tab:choicesTested}. We will later show how they influence the detection of epileptic seizures. 

    \begin{table}
    \caption{Overview of all methodological choices tested} 
    \label{tab:choicesTested}
    \renewcommand{\arraystretch}{1.2}%
    \begin{tabular} {p{16mm}p{60mm}}
        % 		\begin{tabular}{ccccc}
        \textbf{Data used} & Subsets of data with different imbalance ratios (e.g. Factor1 and Factor10)\\
        \textbf{} & All data, with different splits into training folds (SeizureToSeizure, 1h/4h windows)\\
        \hline
        \textbf{Training } & Cross-validation type: Leave-one-out (L1O) or Time-series-CV (TSCV)\\
        \textbf{} & Window step: 0.5 to 4s, with 4s windows\\
        \textbf{} & Personalized models or generalized models \\
        \hline
        \textbf{Perf. } & Episode and duration-level performance \\
        \textbf{metric} & Micro or macro CV folds averaging \\
    \end{tabular}
    % 		\vspace{5cm}
	\end{table}
    
    \vspace{-2mm}
	\subsection{Data preparation}
	\label{subsec:dataprep}
    \vspace{-2mm}
    An important part of evaluating machine learning algorithms is the data used to train and test the algorithm. A well-known practice is that training, validation, and test subsets must be chosen without overlap and be statistically independent to avoid the effect known as 'data leakage'. But the question that is less discussed is how representative are the data that we use. With the increasing amount of big data collected using the Internet of Things (IoT) and wearable devices, big data sets are no longer rare. Such large datasets are incredibly valuable and essential for having more ML/AI-powered devices in everyday life, but they also bring certain challenges. Training on such a huge amount of data, especially for computationally demanding or memory intensive algorithms or without lots of computational resources, can be complex, slow, and even potentially not feasible. For this reason, a common approach is to create smaller subsets of available datasets. 

    In the case of epilepsy, it is characterized by recurrent but unpredictable electrical discharges in the brain. Epilepsy episodes can last from a few seconds to a few minutes. Overall, when looking at the recorded data, the percentage of seizure data is extremely small, commonly less than 0.5\%. 
    %In Table~\ref{tab:datasets} we show characterization, including the percentage of seizure data, for two publicly available datasets that contain completely continuous recordings recorded from several days (rather than preselected subselection of data, that is common in other datasets). 
    This huge imbalance in epilepsy recordings leads to the common choice of creating a data subset that contains all seizure signals but only a reduced amount of non-seizure signals. This step of generating smaller or even balanced datasets makes training simpler, performance reporting clearer, and speeds up the research process. Most papers tackling the epilepsy detection problem do not use whole long-term epilepsy recordings, but rather data subsets and also very rarely discuss the influence of this decision on their results. 

	% \begin{table}
	% 	\caption{Characterization of two publicly available long-term epilepsy databases (NrSbj: number of subjects, NrS: number of seizures, DatLen: dataset length [h],  SLen: total seizure duration [h], PSLen: percentage of seizure duration [\%])} 
	% 	\label{tab:datasets}
	% 	\renewcommand{\arraystretch}{1.2}%
	% 	\begin{tabular} {p{22mm}p{8mm}p{6mm}p{8mm}p{8mm}p{10mm}}
	% 		% 		\begin{tabular}{ccccc}
	% 		\textbf{Dataset} & \textbf{NrSbj} & \textbf{NrS} & \textbf{DatLen} & \textbf{TSLen}  & \textbf{PSLen}\\
	% 		\hline
	% 		\textbf{CHB-MIT}    & 24 & 198 & 982.9h & 3.2h & 0.32 \% \\ 
 %            \textbf{SWEC-ETHZ}  & 18 & 116 & 2656h & 3.6h & 0.13 \% \\ 
	% 	\end{tabular}
	% 	% 		\vspace{5cm}
	% \end{table}

    In this paper, we address this question by testing several epilepsy subsets created from a main dataset. We evaluate the influence of using all or only some data samples, as well as the impact of the seizure to non-seizure imbalance ratio. We also test the influence of data splitting during the training and cross-validation folds and show that this choice can be very critical and make a big difference in whether the proposed algorithm will work in practice when all data are used, without the possibility of performing any selection. 
    
    \begin{figure}
		\centering
		\includegraphics[width=1\linewidth]{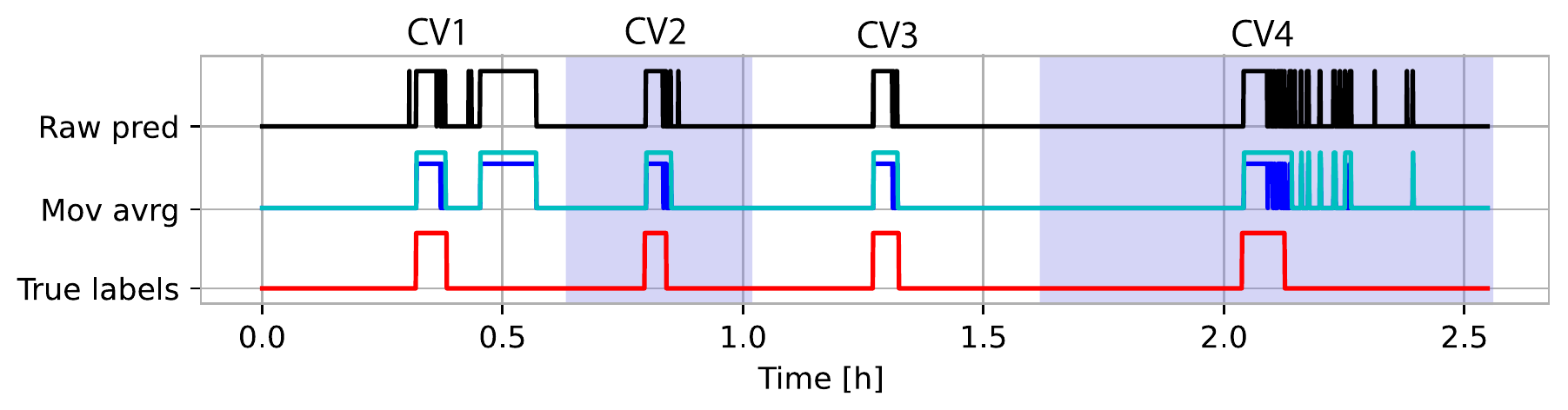}
        \vspace{-5mm}
		\caption{\small{Epilepsy model predictions example. Predictions without any post-processing and with two types of post-processing as well as true labels are shown. Of interest are the distributions of false positives. The distributions of false positives are of particular interest.} } 
		\label{fig:dataPredictionExample}
	\end{figure}

    \vspace{-2mm}
    \subsection{Generalized vs. personalized models}
	\label{subsec:genPersmodels}
    \vspace{-2mm}
    In many applications where underlying data patterns are highly specific, such as in many biomedical use cases, there are two approaches to training; personalized and generalized training and models. Epilepsy is a good example of this, where underlying electroencephalography (EEG) patterns are highly variable between EEG channels, recording sessions, and subjects. Personalized training means that data from the same subject are used to train the model. %In practice, it means that training is usually done either in leave-one-seizure/file-out or TSCV approach, and then performance is accumulated for all seizure tests to be reported as the final result. 
    This leads to as many ML models as subjects we have. 
    Generalized training, on the other hand, would lead to a single ML model for all subjects. To avoid data leakage (and enable comparison with personalized models), every subject has its own generalized model trained on all subject's data but that test subject, which is also known as the leave-one-subject-out approach. 

    On one hand, personalized modes can capture subject-specific patterns better, but are also trained on less data in total, which can sometimes be limiting, as some subjects have very few seizures recorded. 
    %(PUT MAYBE DISTIBUTION OF NUMBER OF SEIZURES IN CHBMIT)  - no space
    On the other hand, generalized models are more complex to train as they are trained on more data, and can also be less subject-specific but may be more interesting for building large-scale wearable outpatient systems. 
    %For widely accessible wearable devices, it will not be realistic to record each subject's data to create personalized models, but generalized models will have to be used. This is also necessary for subjects that have very rare occurrences of seizures, which cannot be captured in a few days of in-hospital recordings.
    
   \begin{figure}
		\centering
		\includegraphics[width=1\linewidth]{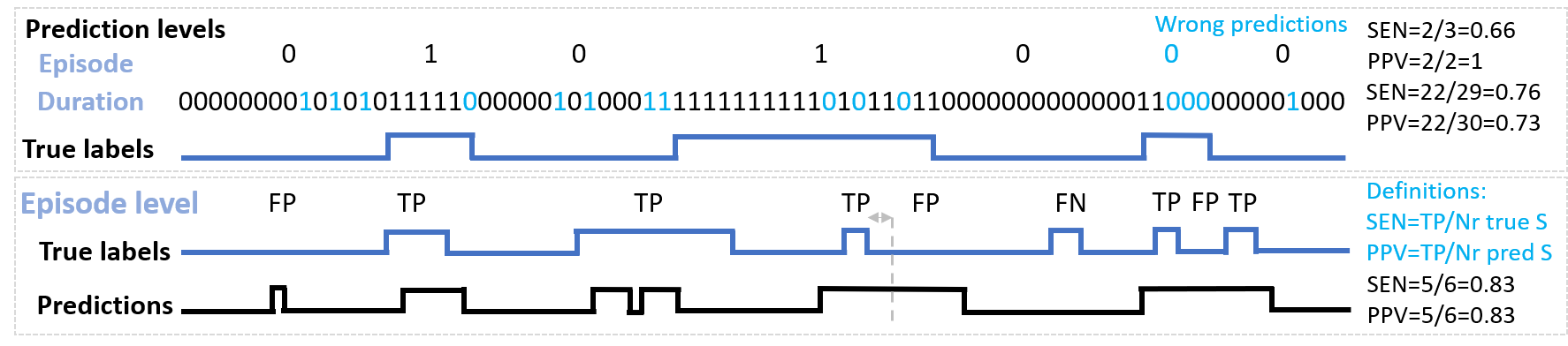}
        \vspace{-2mm}
		\caption{\small{Illustration of duration and episode-based performance metrics. }} 
		\label{fig:perfMetricsIllustration}
	\end{figure} 
 
	\subsection{Respecting temporal data dependencies}
	\label{subsec:temporaldata}
    \vspace{-2mm}
    Another aspect of data that is commonly forgotten is that all data are recorded in time and that sometimes this imposes some unavoidable statistical dependencies. Some underlying patterns that our ML algorithms can use can only exist in a certain order, and for this reason it might not be fair to use data that are in the future to train and then test using data that was before it. On one hand, it can miss some patterns useful for detection, and on the other hand, it can lead to potentially unfeasible results for in-practice applications. 

    Two parts of the ML workflow must be considered in this case. Often, data samples are shuffled before training, whereas for temporal data, this might not be advisable. Furthermore, if some statistical knowledge on the distribution and length of certain classes is available, this knowledge can be used to post-process predicted labels and lower misclassification chances. For example, in the case of epileptic seizures, it would not be realistic. 
    %that there are seizures that last only a few seconds (<5s) or if seizures are too close (<1min REF??), they are probably part of the same seizure.
    that an individual suffers an epileptic seizure of 1-second duration every minute.
    % This statistical knowledge can be used to post-process labels and report more realistic performance if time information is kept and data is not shuffled before training or testing. 

    Second, the temporal aspect of the data is relevant when choosing the cross-validation (CV) approach. A common CV approach for personalized training is leave-one-seizure-out, which means that data from one seizure is left out for testing, and seizures that come before but also after will be used for training. 
    On the other hand, in the time series cross-validation (TSCV) approach~\cite{mohammed_time-series_2021, zanetti_approximate_2022} only previously acquired data can be used for training. This means that if files are ordered in time, for the first CV fold only one file will be used for training and the one after it for testing. For the following CV folds, one more file is always added to the training set (the file previously used for testing), and testing is done on the next available file. This CV approach is rarely used in the literature but is the only feasible approach for online data training (and inference) on wearable devices.
    % For the second CV, two files (the first on and the one in which previously was testing) will be used for training and the third file for testing, and so on.

    \subsection{Data segmentation}
	\label{subsec:dataPartitioning}
    \vspace{-2mm}
    Typically, features are extracted from fixed-size windows of data and calculated with the'moving window' repeatedly in shifts of a chosen step size. Here, two parameters have to be decided: window size (WS) and window step size (WSS) for which we move the feature extraction window. Choosing a larger window size might be necessary when extracting frequency information, but it also limits the possibility of detecting very short patterns. Similarly, a smaller step size can decrease detection latency, but increases the computational costs of the algorithm due to more frequent feature extraction. These parameters can be optimized according to several aspects: features used and their properties and complexity, latency requirements, or available computational resources. If none of these is limiting, the parameters are generally optimized in terms of performance. It is interesting to note how much performance can change depending on these choices. More importantly, parameter choice and the reasoning behind it should be mentioned and documented in papers.

    \subsection{Evaluation metrics}
	\label{subsec:EvalMetrics}
    \vspace{-2mm}
    For temporal and sequential data, standard performance evaluation metrics, such as sensitivity and specificity, may not always be the most appropriate and can even be misleading~\cite{japkowicz_evaluating_2011}. Evaluation metrics must ultimately reflect the needs of users and also be sufficiently sensitive to guide algorithm development~\cite{ziyabari_objective_2019}.
    As Shah et al. stated in~\cite{shah_validation_2020} there is a lack of standardization in the evaluation of sequential decoding systems in the bioengineering community. 
    
    The same authors compare five popular scoring metrics for sequential data in~\cite{ziyabari_objective_2019}. Among them, the most interesting are 'Epoch-based sampling' (EPOCH), 'Any-overlap' (OVLP), and 'Time-aligned event scoring' (TAES). 
    EPOCH treats the reference and hypothesis as temporal signals, samples them at a fixed epoch duration, and counts errors (TP, TN, FP, and FN) accordingly. For an epoch duration of one sample, this metric processes data sample-by-sample and results in typical performance measures (such as accuracy, sensitivity, specificity, F1 score etc).  
    The OVLP measure~\cite{ wilson_seizure_2003}, interprets signals as a series of same-label episodes and then assesses the overlap in time between reference and hypothesis. It counts a 'hit' in case there is any overlap between the reference and hypothesis. In Fig.~\ref{fig:perfMetricsIllustration} we illustrate several use cases, how errors are counted, and what is the final performance measure. 
    % This performance metric is common in speech ad image recognition applications under the name term-base performance~\cite{fiscus_results_nodate}.
    The authors in~\cite{ziyabari_objective_2019} also propose the TAES metric which combines the EPOCH and OVLP information into one metric. The approach is very similar to OVLP, but rather than simply considering if there is any overlap between reference and hypothesis episodes, the percentage of overlap is measured and weighs the errors (TP, TN, FP, FN) accordingly. 
    Here, we want to demonstrate the difference in performance in the use case of epilepsy detection, depending on the chosen performance measure, and also how these performance metrics can be used to interpret the quality of algorithm predictions. The code of those metrics is available online\footnote{https://c4science.ch/source/PerformanceMetricsLib/}. 

    Another performance measure with a strong practical impact, and thus often used for epilepsy detection, is the false alarm rate (FAR), or the number of false positives per hour/day. Clinicians and patients see this measure as more meaningful than many more commonly used metrics, and are very demanding in terms of performance, requiring it to be as low as possible for potential wearable applications (e.g., less than 1 FP/day)~\cite{shah_validation_2020}. This also necessitates exceptionally high constraints on the required precision (usually much higher than 99\%). 
 
    Finally, to quantify global performance, the accumulated performance of all cross-validation folds has to be calculated. But here are also choices to be made. One can measure the average performance of all CV folds (micro-averaging) or can, for example, append predictions of all test files next to each other and only then measure performance on all appended data (macro-averaging). In Fig.~\ref{fig:dataPredictionExample}, an example of predictions (also with moving average post-processing) is given for all files of one subject. What is important to notice is the distribution of false positives over time and over the different files/CV folds. Most often, false positives occur around seizures. However, there can potentially be a fold(s) with an unexpectedly large number of false positives. If the final performance is measured as an average of the performances of each fold, a fold with many false positives, as in Fig.~\ref{fig:dataPredictionExample}, will have a lower influence on the total performance than if all predictions are appended and performance is measured only afterward. This potential overestimation of performance when averaging cross-validations should also be taken into account. 

    \begin{figure}
		\centering
		\includegraphics[width=0.8\linewidth]{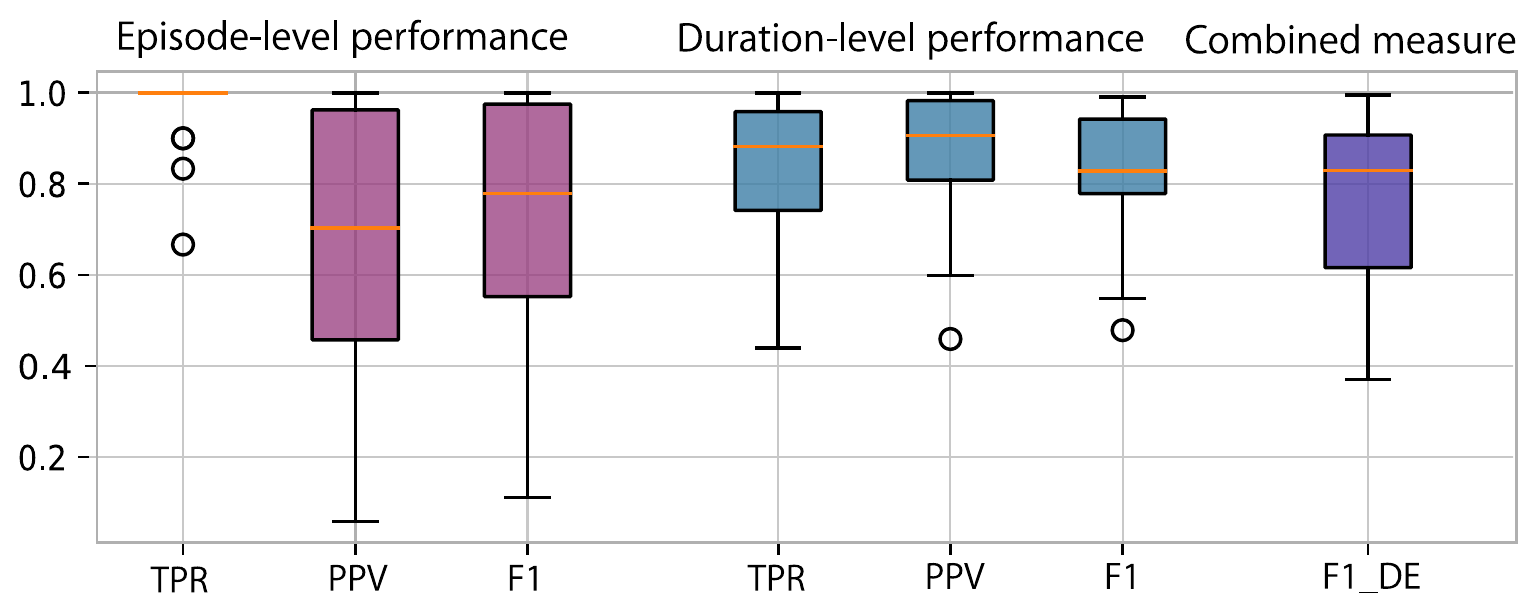}
        \vspace{-2mm}
		\caption{\small{Epilepsy detection performance measured through seven measures; both on episode and duration level. For each, sensitivity (TPR), precision (PPV) and F1 score are measured. The results show average performance for all 24 subjects for 'Fact1' data subset.}} 
		\label{fig:perfMetricsRes}
	\end{figure}

	%%%%%%%%%%%%%%%%%%%%%%%%%%%%%%%%%%%%%%%%%%%%%%%%%%%%%%%%%%%%%%%%%%%%%%%%%%%%%%%%%%%%%%%%%%%%%%%%%%%%%%%%%%%%%%%%%%%%%%%%%%%%%%%%%%%%%%%%%%%%%%%%
    \vspace{-2mm}
	\section{Experimental setup}
	\label{sec:Expsetup}
    \vspace{-2mm}
    \subsection{Dataset}
	\label{subsec:Dataset}
    \vspace{-2mm}
    In this work, we use the CHB-MIT epilepsy dataset, an open source widely used dataset for the detection of epilepsy~\cite{shoeb_application_2009}, as it is a good representative of continuous, relatively long-term monitoring (over several days).
    CHB-MIT is an EEG database, with a total of 982.9 hours of data recorded at 256Hz. It consists of 183 seizures forming a total of 3.2 hours or 0.32\% of labeled ictal data, from 24 subjects with medically resistant seizures ranging in age from 1.5 to 22 years. On average, it has 7.6 $\pm$ 5.8 seizures per subject, with an average seizure length of 58.6 ± 65.0 s. It was recorded using the bipolar montage (10-20 system) and thus contains between 23 and 26 channels, of which we use the 18 channels that are common to all patients\cmmnt{ (i.e., FP1-F7, F7-T7, T7-P7, P7-O1, FP1-F3, F3-C3, C3-P3, P3-O1, FP2-F4, F4-C4, C4-P4, P4-O2, FP2-F8, F8-T8, T8-P8, P8-O2, FZ-CZ, CZ-PZ)}.
		
    \subsection{Machine learning training}
	\label{subsec:MLmodel}
    \vspace{-2mm}
 	We extract 19 features from each of the 18 channels, similar to~\cite{pale_hyperdimensional_2022}, calculating them on 4-second windows with a moving step of 0.5 seconds (unless otherwise specified). We use two time-domain features, mean amplitude and line length, and 17 frequency domain features. Both relative and absolute values of power spectral density in the five common brain wave frequency bands are used: delta: [0.5-4] Hz, theta: [4-8] Hz, alpha: [8-12] Hz, beta: [12-30] Hz, gamma: [30-45] Hz, and low-frequency components: [0-0.5] Hz and [0.1-0.5] Hz.
    Before extracting the features, the data is filtered with a 4th-order, zero-phase Butterworth bandpass filter between [1, 20] Hz. 

    As an algorithm to test the range of parameters mentioned, we choose a highly popular but also feasible algorithm for wearable and outpatient monitoring devices. We implemented a random forest classification algorithm, which is based on an ensemble of 100 decision trees to reduce model overfitting. It is fast and lightweight, both in model size and memory footprint~\cite{zanetti_real-time_2022}, and has been used extensively for EEG-based seizure classification~\cite{siddiqui_review_2020, sopic_e-glass_2018, zanetti_robust_2020}.
    In the end we postprocess predicted labels with a moving average window of 5 seconds, and majority voting to smooth predictions and remove unrealistically small seizures. If seizures are closer than 30s, we merge them into one.

	\begin{figure}
		\centering
		\includegraphics[width=0.95\linewidth]{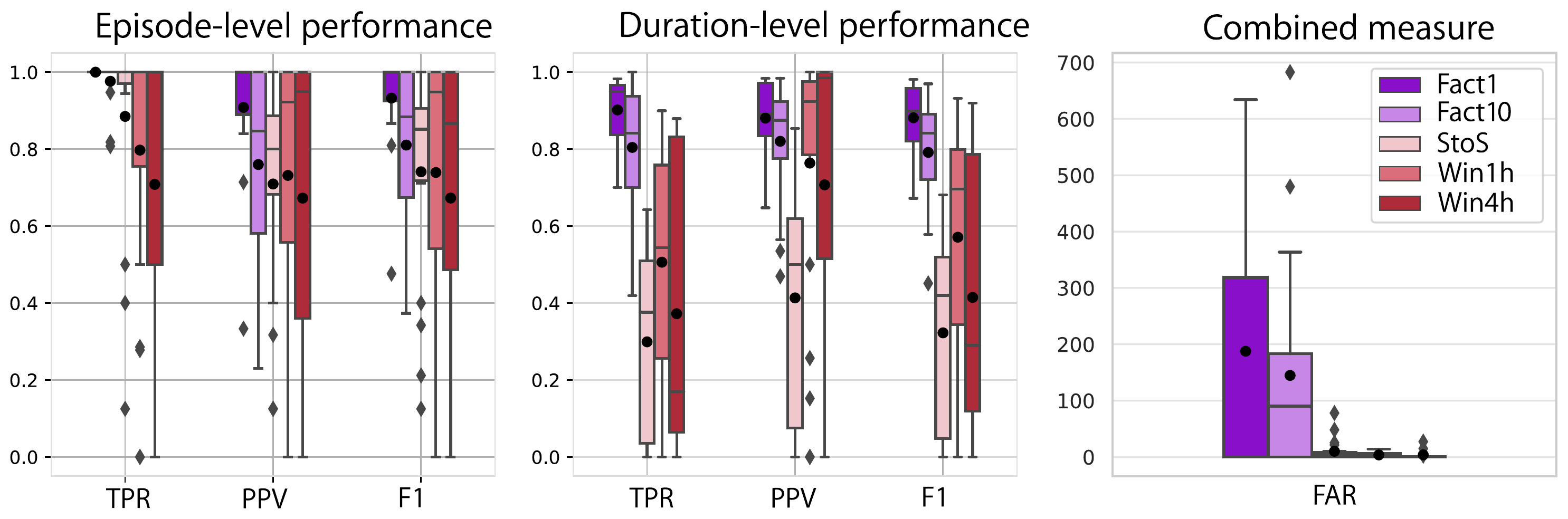}
        \vspace{-2mm}
		\caption{\small{Epilepsy detection results depending on the data subset used.}} 
		\label{fig:datasubsetRes}
	\end{figure}

    %%%%%%%%%%%%%%%%%%%%%%%%%%%%%%%%%%%%%%%%%%%%%%%%%%%%%%%%%%%%%%%%%%%%%%%%%%%%%%%%%%%%%%%%%%%%%%%%%%%%%%%%%%%%%%%%%%%%%%%%%%%%%%%%%%%%%%%%%%%%%%%%
    \vspace{-2mm}
    \section{Experimental results}
	\label{sec:ExpResults}
    \vspace{-2mm}
    \subsection{Evaluation metrics}
	\label{subsec:EvalMetricsRes}
    \vspace{-2mm}
    In this work, we use two metrics to measure performance. One is the EPOCH approach described in Sec.~\ref{subsec:EvalMetrics}, with an epoch duration of one sample. TP, TN, FP, and FN are detected sample by sample, which are further used to calculate the sensitivity (TPR), precision (PPV), and F1 score. We call this duration-based performance, which characterizes how well seizures are detected with respect to their full length. 
    The other performance metric is OVLP, which detects overlaps between predicted (hypothetic)  and reference seizure or non-seizure episodes. We call this an episode-based metric, as it cares only about whether each of the seizure episodes has been detected, not caring exactly about the predicted seizure duration.
    These two metrics are very easily interpretable. For example, if the sensitivity at the episode level is 80\% and there were 10 seizures, it means that 8 episodes were detected but 2 were missed. 
    The TAES metric proposed in~\cite{ziyabari_objective_2019} is an interesting approach to combine both metrics but is harder to interpret, and thus it is not used here. 

    Fig.~\ref{fig:perfMetricsRes} shows the average performance of personalized models for the 24 subjects from the balanced CHB-MIT dataset. For both episode- and duration-based performance, sensitivity, precision, and F1 scores are shown, as well as one accumulative measure, the mean of F1 score for episode and duration-level ('F1\_DE'). The sensitivity of the episode is on average 100\%, which means that except for a few cases, all episodes of seizures were detected. Looking at duration-level sensitivity, it is clear that even if seizure episodes were perfectly detected, their whole duration was not always predicted. Looking at the precision, it is clear that there are also false positive predictions, and more of them when measuring on episode level than duration level, meaning that there were many short false positives. Observing the performance through these six values enables a more complete characterization of the prediction performance of a certain algorithm. It also enables a more nuanced comparison between different methodological steps or parameter values used, as will be shown next.

 	\begin{figure}
		\centering
		\includegraphics[width=0.95\linewidth]{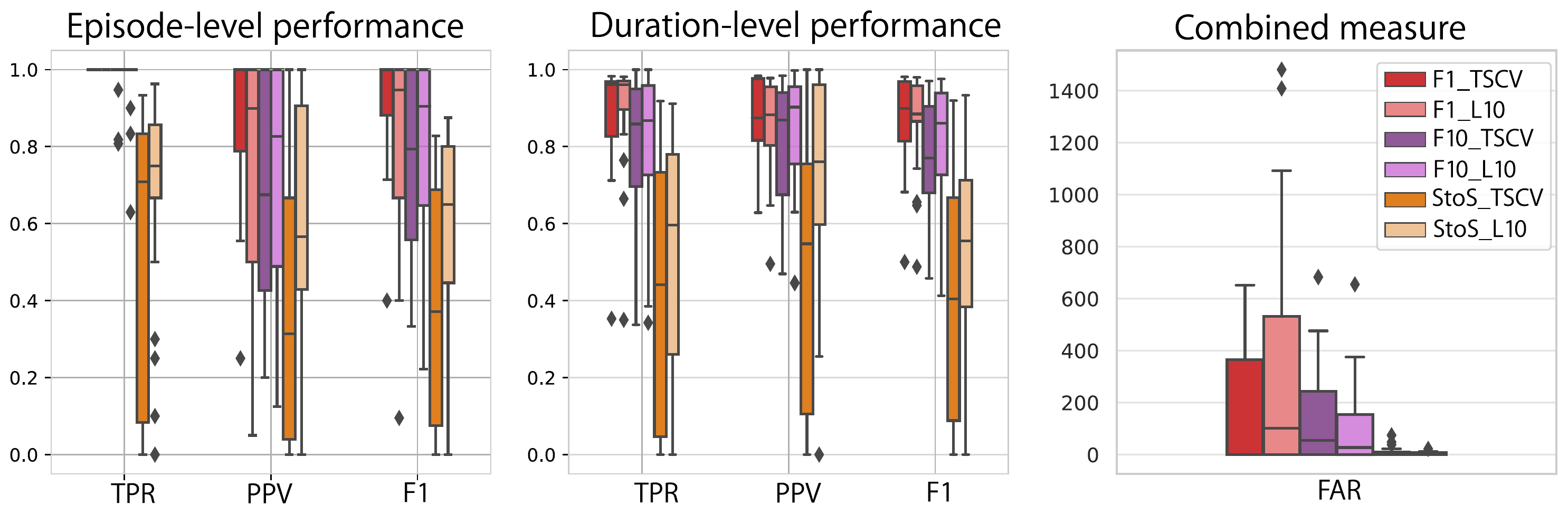}
        \vspace{-2mm}
		\caption{\small{Performance results when using leave-one-out vs. time-series cross-validation. Results are shown for three data subsets (F1, F10 and StoS). }} 
		\label{fig:L1OvsRBres}
	\end{figure}
 
    \vspace{-2mm}
    \subsection{Generalized vs. personalized models}
	\label{subsec:genPersmodelsRes}
\begin{figure}
		\centering
		\includegraphics[width=0.95\linewidth]{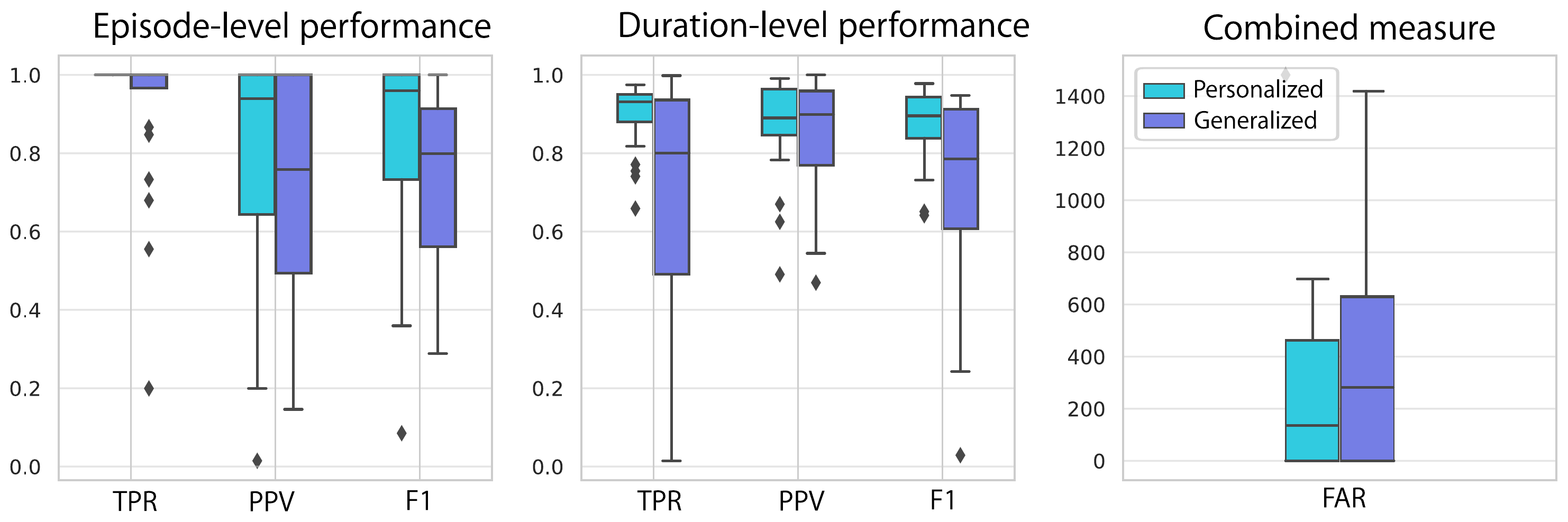}
        \vspace{-2mm}
		\caption{\small{Comparison of average performance (for all subjects) of personalized and generalized models.}} 
		\label{fig:PersGenRes}
	\end{figure}

    \vspace{-2mm}
    Here we test the performance difference when training both personalized and generalized models for each subject. We used a balanced data subset ('Fact1'). In Fig.~\ref{fig:PersGenRes}, the average for all subjects is shown, making clear the lower performance of generalized models. Inspecting the performance of the generalized model per subject reveals a clear distinction between patients on which generalized models perform very well and those for whom it performed poorly, either due to many false positives or almost no detected seizures. How to create generalized models and whether there should be subtypes of generalized models for different patient groups remains a question for future generations. For the remainder of this work, we will focus on personalized models. 
 
    \subsection{Data preparation}
	\label{subsec:dataprepRes}
    \vspace{-2mm}
    In Fig.~\ref{fig:datasubsetRes}, the performance of epilepsy detection is shown for five different data subsets used to train and test. The first two approaches, 'Fact1', and 'Fact10', contain a subset of the original CHB-MIT dataset in two different ratios of seizure and non-seizure data. 'Fact1' is a balanced data subset that has the same amount of seizure and non-seizure data, where all available seizure data are used, along with a randomly selected equal amount of non-seizure data. The 'Fact10' subset is constructed similarly, with the difference that the amount of randomly selected non-seizure data is 10x more than seizure data. Data are divided into files equal to the number of seizures with one seizure per file. Each file is arranged such that seizure data occur in the middle of the file, with nonseizure data split on both sides. Therefore, the total file length depends on the length of the seizure and the factor value. This organization enables easier training in the case of the leave-one-seizure-out approach, as each file is equally balanced. 

    The last three approaches, Seizure To Seizure ('StoS'), and 1 or 4 hour windows ('Win1h'/'Win4h', 'WinXh' together), contain all data samples from the CHB-MIT database but are rearranged into files containing different amounts of data. The 'StoS' approach consists of files that start with the non-seizure data after the previous seizure and end when the next seizure ends. In this way, every file contains exactly one seizure, but the entire length of the file is not fixed. The last two approaches, 'Win1h' and 'Win4h', as the names imply, divide the dataset into files of 1-hour or 4-hour duration. In this way, some of the files may contain zero to possibly multiple seizures. In all three cases, we trained using time-series cross-validation. We specified that the first file must contain a certain amount of data (five hours) and at least one seizure, and as such it is slightly different from other consequent files.

    We analyze the impact of data preparation considering three metrics: false alarm rate (FAR), sensitivity (TPR), and precision (PPV). First, the false alarm rate, defined here as the number of FPs per day, as shown in Fig.~\ref{fig:datasubsetRes}, is significantly higher for data subsets (Fact1(0)) than for the whole dataset training (StoS or Win1h/4h). This can be traced to two reasons. The first is that Fact1(0) is trained on much less non-seizure data, potentially not enough to model properly the non-seizure patterns, resulting in more non-seizure samples being falsely classified. The second is that because testing is done only on a subset of data, FPs must be linearly scaled to estimate the false alarm rate per day, potentially leading to very high numbers. For these reasons, data subsets should not be used to estimate the false alarm rate of an epilepsy detection algorithm.

    Next, we consider the sensitivity and precision. For Fact1(0), seizure episode detection is easier, visible from episode-level sensitivity that is 100\%. Detecting all seizure episodes perfectly is much harder when the entire dataset (StoS or WinXh) is used and visible by episode sensitivity values ranging between 80 and 95\%. Precision presents a more complex picture, dropping sharply for StoS before recovering for WinXh. This is because StoS retraining and testing only occurs after every new seizure, where in the meanwhile could have been hours of non-seizure data.  The WinXh strategies retrain much more often, thus making it easier to learn non-seizure patterns and lower false predictions. To conclude, using all data significantly reduces false positives, but also results in lower, but more realistic, sensitivity and precision values. 
    %When looking at the performance results (average for all subjects) we can notice a significant difference in performance, as shown in Fig.~\ref{fig:datasubsetRes}. First, if we look only at the number of FPs per day, data subsets ('Fact1' and 'Fact10') have an extremely large amount of FPs. Due to the small amount of non-seizure data this estimation is not realistic, and thus it is not advised to use data subsets for a fair estimation of the false alarm rate. 
    %Next, sensitivity on the level of episodes is 100\%, but is lower when all data is used and ranges between 80 and 95\%. When looking at precision, as the amount of non-seizure data is increased, performance is lower due to more possibilities for false positives. For this reason, results reported on data subsets are potentially very overestimated. Here we see also that 'StoS' approach seems to be, in general, the most demanding as sometimes there can be very long times before the next seizure occurs, and it is useful to retrain with the new data in the meanwhile, too. For this reason, using fixed-size windows is more appropriate. To conclude, using all data significantly reduces false positives but also results in lower, but more realistic sensitivity values.

    \subsection{Respecting temporal data dependencies}
	\label{subsec:temporaldataRes}
    \vspace{-2mm}
    To show the influence of cross-validation choices on performance results, we trained and tested three data subsets with different data imbalance ratios ('Fact1', 'Fact10', 'StoS'), both using leave-one-seizure-out (L1O), and time-series cross-validation (TSCV) approaches. The results are shown in Fig.~\ref{fig:L1OvsRBres}. The superior performance of the L1O approach is evident in almost all aspects (sensitivity, precision, and numFP). This is reasonable as more data were used to train with L1O than with TSCV approach. The difference in performance ranged from 3 to 7\% for the F1 score in episode detection. This is not a recommendation to use L1O; in contrast, it demonstrates that training on future data leads to overestimated performance and should be avoided. 

    \vspace{-2mm}
    \subsection{Data segmentation}
	\label{subsec:dataPartitioningRes}
\begin{figure}
		\centering
		\includegraphics[width=0.95\linewidth]{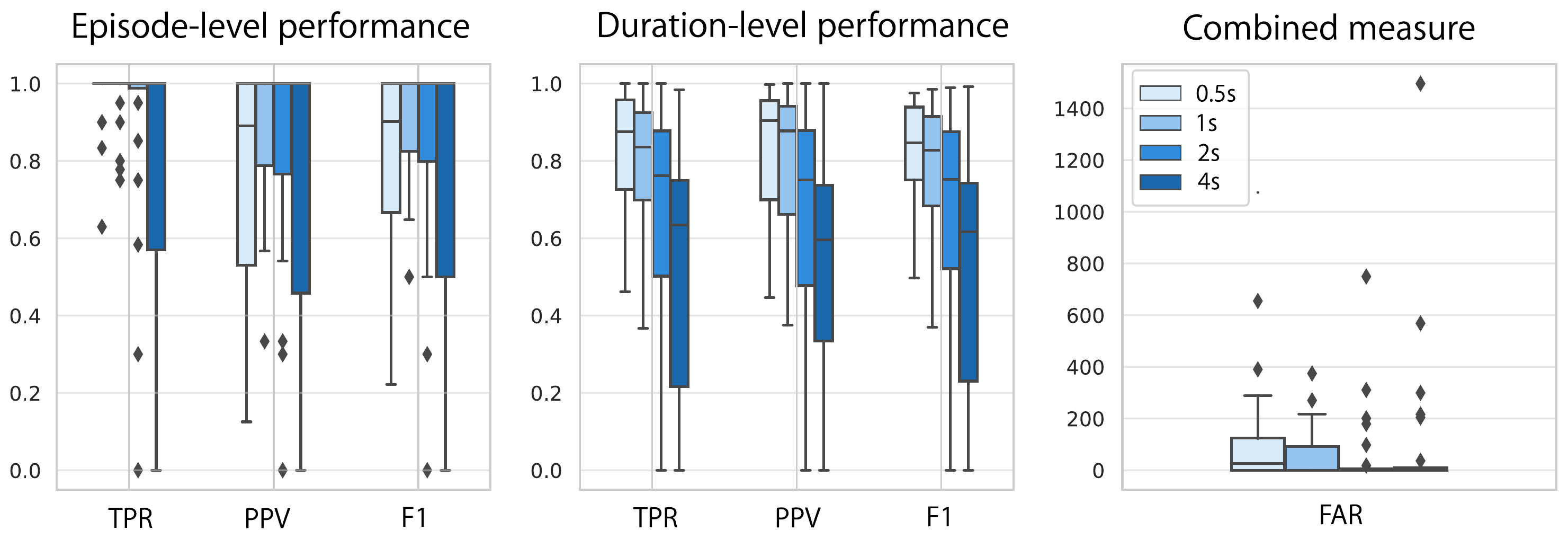}
        \vspace{-2mm}
		\caption{\small{Performance with respect to different window step sizes. Steps of 0.5, 1, 2 and 4s were used, with window size of 4s. }} 
		\label{fig:windowSizeRes}
	\end{figure}

    \vspace{-2mm}
    In Fig.~\ref{fig:windowSizeRes}, the performance of epilepsy detection is shown when the window step ranged from 0.5 to 4 s, with a window size of 4 s. The most clear and expected pattern is that by increasing the step size, the number of false positives is reduced significantly, but what is interesting is that the proportion of FP increases, thus reducing the precision for episodes. More precisely, precision increases first (while sensitivity is still high) and then drops for episodes (because less seizures were also detected), leading to the conclusion that too big steps are risky. 
    Increasing window step size also reduces sensitivity, more noticeably for duration level, which may be due to the fact that with large steps, shorter seizures can be potentially missed. 
    These results demonstrate the complexity of the window step size parameter, that it is beneficial to experiment before choosing one value, and that it has to be necessarily reported to make the results comparable and reproducible. 

    %%%%%%%%%%%%%%%%%%%%%%%%%%%%%%%%%%%%%%%%%%%%%%%%%%%%%%%%%%%%%%%%%%%%%%%%%%%%%%%%%%%%%%%%%%%%%%%%%%%%%%%%%%%%%%%%%%%%%%%%%%%%%%%%%%%%%%%%%%%%%%%%
    \vspace{-2mm}
    \section{Discussion}
	\label{sec:Discusion}
    \vspace{-2mm}
    \subsection{Data aspects}
	\label{subsec:dataAspecs }
    \vspace{-2mm}
    As seen from results in Sec.~\ref{subsec:dataprepRes}, using all the data significantly reduces false positives and also results in lower and more realistic sensitivity values. Thus, if computational and memory resources are sufficient, models should be trained using all available data. If this is not possible, then the subset of data should contain significantly more non-seizure data. %(at least 10x).
    However, data subsets will result in an unrealistic false alarm rate. When using the whole dataset, the most appropriate seems to be to use fixed-size time frames in which models are retrained/updated regularly. The size of this time frame should also be tested and reported. General advice would be to use data subsets for initial experimentation and building an understanding of the algorithm and its parameters, but that all the available data should be used for reporting the final performance. It is also useful to take into account and characterize the class imbalance.

    % Other interesting aspect that is worth mentioning and is often used, is data augmentation. Especially in cases with highly imbalanced datasets, such as epilepsy datasets, it is tempting to augment seizure data. For example, using random oversampling (ROS) or 'synthetic minority oversampling' (SMOTE) can lead to improved performance results.  

    % Similarly, for epilepsy datasets sometimes researchers exclude pre- or post-seizure data. Determining the exact seizure start and ending moment is quite demanding, and often, two neurologists will have a disagreement in a few seconds~\cite{nascimento_one_2022}. For this reason, excluding pre- and post-ictal data can be tempting. This, again often leads to better performance due to easier detection tasks but is also not realistic for real-life applications, and from this reason we do not analyse it her, nor recommend to use in practice. 

    When talking about the temporal aspect of data, several things should be taken into account. We advise not to shuffle data samples before training and testing but rather to use temporal information and knowledge on class distribution to postprocess predicted labels, which can increase performance, but must also be clearly reported. 
    
    Finally, it is critical to decide whether to use only the data from the same subject to create personalized and, as shown, more precise models or to use all available data from other subjects to create generalized models. Generalized models can have lower individual performance, but can be used for new subjects. This topic represents a full research topic on its own.
    For example, future research should investigate whether it is possible to create models that can use generalized models as a starting point from which they can be personalized.Would these models lead to better overall performance in comparison to personalized models? Would less personal data be needed to personalize models if generalized modes are used as starting point? 
    Similarly, the unavoidable question is, can we somehow profit from both generalized and personalized models? Can we combine them in some beneficial way? 
    
    \vspace{-2mm}
    \subsection{Training aspects}
	\label{subsec:trainAspecs }
    \vspace{-2mm}
    When talking about the choice of cross-validation as shown in Sect.~\ref{subsec:temporaldataRes}, the leave-one-out approach leads to higher performance than if data are trained in temporal order using time series cross-validation. However, the L1O approach is not realistic for training data in real time, while TSCV is intended for such scenarios. Training models online as data are being acquired is one of the necessary next steps for ML models on IoT devices, and thus TSCV will have to become the standard method. 

    Data partitioning has two parameters that can also play a significant role in performance, namely the window size used to extract features and the window step size. Their optimal choice can depend on each use case, the features extracted and their properties and complexity, latency requirements, and available computational resources. Here, we show how window step size can influence performance, with different patterns for false alarms, sensitivity, and precision, and how it has different impacts on duration- or episode-level classification. The results show the complexity of the window step size parameter, indicating that it is beneficial to test it before choosing one value and that it must be reported to make results comparable with the literature. One research avenue that we have not considered here as it is outside the scope of this work, but which can be potentially very beneficial, is optimizing the window size parameter for each feature individually. 

    \vspace{-2mm}
    \subsection{Performance estimation aspects}
	\label{subsec:performanceAspecs }
    \vspace{-2mm}
    Here we proposed to use two performance metrics, one at the duration-level and one at the episode-level. Each of them has certain advantages, and thus their values should be interpreted carefully. Nevertheless, together they provide a full picture of the detection characteristics of the algorithm analyzed. 
    
    For example, EPOCH, a duration-based metric, cares about the duration of the events and thus weighs long events more importantly. This means that if a signal contains one very long seizure event and some shorter ones, the accuracy with which the long event is detected will dominate the overall scoring. In epilepsy detection, as in many applications, the duration of the event can vary dramatically; therefore, this must be taken into account. 
    For this reason, OVLP, an episode-level performance metric, is much easier to interpret. However, such an episode-level metric is more permissive and tends to produce much higher sensitivities. It can also be implemented so that if an event is detected in close proximity to the reference annotation, it is considered correctly detected, which can further increase the performance values. 
    %This is not necessarily negative, as, for example, if the seizure is detected within a certain time (e.g., 30s) before a seizure, it can also be interpreted as seizure prediction. On the other side, if it is also detected after seizures, it could be that it is detecting inter-ictal period or post-ictal epilepsy discharges, and this detection should not necessarily be treated as a false positive. 
    
    Nowadays, in the literature, duration-level-based performance is still the most popular, but there are trends of moving toward more event/episode-based performance measures~\cite{ziyabari_objective_2019}.
    %For example recent challenge for epilepsy detection~\cite{Epilepsy_challenge} requires using both duration and episode based metric. 
    Currently, there is no standardization. Until then, the performance metrics used, as well as post-processing that has been utilized to smooth the labels, must be clearly described.  

    Similarly, the method to achieve the overall performance measure from the individual CV folds must be documented. We recommend that overall performance be calculated by temporally appending all fold predictions in time, rather than as the average of all fold performances. %This is due to potential specific patterns in the distribution of false positives (such as all in one file) that can lead to overestimated performance when reporting the average. 
    For example, if one CV fold (or small portion, as in Fig.~ref{fig:dataPredictionExample}) has extremely high number of false positives, but all other ones have good performance, it will affect overall estimation of e.g., precision much less (due to averaging over all folds), leading to potentially overestimated performance.

	%%%%%%%%%%%%%%%%%%%%%%%%%%%%%%%%%%%%%%%%%%%%%%%%%%%%%%%%%%%%%%%%%%%%%%%%%%%%%%%%%%%%%%%%%%%%%%%%%%%%%%%%%%%%%%%%%%%%%%%%%%%%%%%%%%%%%%%%%%%%%%%%
    \vspace{-2mm}
    \section{Conclusion}
	\label{sec:Conclusion}
    \vspace{-2mm}
    In this work, we have characterized the influence of a wide range of important methodological choices for epilepsy detection systems. When choosing a subset of the dataset for training, performance can be highly overestimated compared to training in the entire long-term data set. Thus, for real-life performance estimation, using all long-term data is necessary. Similarly, using the leave-one-seizure-out cross-validation approach can improve detection performance, but it is not realistic for online data training, as it uses future data. Thus, we recommend using a time-series cross-validation approach instead, with macro-averaging rather than microaveraging. Training on a generalized level can be challenging due to its subject-specific nature, leaving personalized models outperforming generalized ones. Furthermore, performance metrics must reflect users' needs and be sufficiently sensitive to guide algorithm development. Consequently, we encourage the usage of both episode-based and duration-based performance metrics, which together can give a more nuanced picture of algorithm performance. Finally, whatever choices are made, to further increase the comparability and reproducibility of results, it is essential that all choices and parameters are well reported.

	%\def\IEEEbibitemsep{0pt plus .5pt}
	%\bibliographystyle{bibliographies/myIEEEtran.bst}
		
	%\bibliography{bibliographies/biblio.bib}
	\small
	\printbibliography
\end{document}